\RequirePackage[loading]{tracefnt}
\documentclass[conference]{IEEEtran}
\IEEEoverridecommandlockouts
\usepackage{cite}
\usepackage{amsmath,amssymb,amsfonts}
\usepackage{dsfont}
\usepackage{mathtools}
\usepackage{hyperref}
\usepackage{algpseudocode}
\usepackage{graphicx}
\usepackage{textcomp}
\usepackage{xcolor}
\usepackage{enumitem}
\usepackage{booktabs}
\makeatletter
\let\MYcaption\@makecaption
\makeatother

\usepackage[font=footnotesize]{subcaption}

\makeatletter
\let\@makecaption\MYcaption
\makeatother
\usepackage{tabularx}
\usepackage[separate-uncertainty=true, per-mode=symbol]{siunitx}
\usepackage{pifont}
\newcommand{\cmark}{\ding{51}}%
\newcommand{\xmark}{\ding{55}}%

\usepackage{xcolor,colortbl}
\definecolor{Gray}{gray}{0.85}
\newcolumntype{a}{>{\columncolor{Gray}}c}

\def\BibTeX{{\rm B\kern-.05em{\sc i\kern-.025em b}\kern-.08em
    T\kern-.1667em\lower.7ex\hbox{E}\kern-.125emX}}

\algdef{SE}[SUBALG]{Indent}{EndIndent}{}{\algorithmicend\ }%
\algtext*{Indent}
\algtext*{EndIndent}

\makeatletter
\let\old@ps@headings\ps@headings
\let\old@ps@IEEEtitlepagestyle\ps@IEEEtitlepagestyle
\def\confheader#1{%
	\def\ps@headings{%
		\old@ps@headings%
		\def\@oddhead{\strut\hfill#1\hfill\strut}%
		\def\@evenhead{\strut\hfill#1\hfill\strut}%
	}%
	\def\ps@IEEEtitlepagestyle{%
		\old@ps@IEEEtitlepagestyle%
		\def\@oddhead{\strut\hfill#1\hfill\strut}%
		\def\@evenhead{\strut\hfill#1\hfill\strut}%
	}%
	\ps@headings%
}
\makeatother

\begin{document}

\title{COMET: Contrastive Mean Teacher for Online Source-Free Universal Domain Adaptation}

\author{
	\IEEEauthorblockN{Pascal Schlachter, Bin Yang}
	\IEEEauthorblockA{\textit{Institute of Signal Processing and System Theory}, University of Stuttgart, Germany\\
		\{pascal.schlachter, bin.yang\}@iss.uni-stuttgart.de}
}


\maketitle

\begin{abstract}
In real-world applications, there is often a domain shift (distribution change) from training to test data. This observation recently resulted in the development of test-time adaptation (TTA). It aims to adapt a pre-trained source model to the test data without requiring access to the source data. Thereby, most existing works are so far limited to the closed-set assumption, i.e. there is no category shift (class change) between source and target domain. We argue that in a realistic open-world setting a category shift can appear in addition to a domain shift. This means, individual source classes may not appear in the target domain anymore, samples of new unknown classes may be part of the target domain or even both at the same time. Moreover, in many real-world scenarios the test data is not accessible all at once in form of a dataset but arrives sequentially as a stream of batches which require an immediate prediction. Hence, TTA must be applied in an online manner. To the best of our knowledge, the combination of these aspects, i.e. online source-free universal domain adaptation (online SF-UniDA), has not been studied yet despite its practical relevance. In this paper, we are the first ones to tackle this challenging task. We introduce a Contrastive Mean Teacher (COMET) tailored to this novel scenario. It applies a contrastive loss to rebuild a feature space where the samples of known classes build distinct clusters and the samples of new classes separate well from them. It is complemented by an entropy loss which ensures that the classifier output has a small entropy for samples of known classes and a large entropy for samples of new classes to be easily detected and rejected as unknown. To provide the losses with reliable pseudo labels, they are embedded into a mean teacher (MT) framework. We evaluate our method across two datasets and all category shifts to set an initial benchmark for online SF-UniDA. Thereby, COMET yields state-of-the-art performance and proves to be consistent and robust across a variety of different scenarios. Our code is available at \url{https://github.com/pascalschlachter/COMET}.
\end{abstract}

\begin{IEEEkeywords}
Contrastive Mean Teacher, Online Source-free Universal Domain Adaptation, Online Test-Time Adaptation, Open-World, Open-Set.
\end{IEEEkeywords}

\section{Introduction}
Most machine learning approaches root on the assumption, that the training and test data $(x, y) \in \mathcal{X}\times\mathcal{Y}$ are drawn i.i.d. from the same distribution $\mathcal{P}$ on $\mathcal{X}\times\mathcal{Y}$ \cite{iid_assumption}. Here $\mathcal{X}$ and $\mathcal{Y}$ represent the input and output (or label) space, respectively. However, in real-world applications this assumption is often violated due to various domain (distribution) shifts. In this case, $\mathcal{X}$ and $\mathcal{Y}$ remain the same, but the distribution $\mathcal{P}_t$ of the target domain deviates from the distribution $\mathcal{P}_s$ of the source domain. Accordingly, domain adaptation (DA) to adapt a pre-trained source model to the target domain is required before inference on target data. Most supervised or unsupervised DA approaches assume, in addition to the labeled source dataset, a small labeled or unlabeled target dataset. Recently, a new research area called test-time adaptation (TTA) arose. Its goal is to adapt a pre-trained source model continuously to test-time target domain(s) during deployment of the model by using the unlabeled test-time data only. It can thus be viewed as source-free unsupervised DA. In this way, TTA not only remains applicable when the source data is not accessible due to privacy issues or storage limitations but is also computationally more efficient since it does not reprocess the source data during testing \cite{tent}. However, almost all existing TTA methods for classification are restricted to the closed-set setup, i.e. there is no class change between source and target domain(s).

In realistic open-world scenarios, category shift may happen in addition to domain shift. A category shift denotes the classification situation, where the source label space $\mathcal{Y}_s$ and the target label space $\mathcal{Y}_t$ are different ($\mathcal{Y}_s\neq\mathcal{Y}_t$). There are three main category shifts, namely partial-set ($\mathcal{Y}_t\subset \mathcal{Y}_s$), open-set ($\mathcal{Y}_s\subset \mathcal{Y}_t$) and open-partial-set ($\mathcal{Y}_s\cap\mathcal{Y}_t\neq\emptyset$, $\mathcal{Y}_s\nsubseteq\mathcal{Y}_t$, $\mathcal{Y}_s\nsupseteq\mathcal{Y}_t$). Moreover, typically no prior knowledge about the current category shift is available. Accordingly, a TTA method is required that not only adapts the source model to target distribution(s) but also makes it universally applicable for any kind of category shift by allowing it to reject samples of new classes as unknown while still correctly classifying samples of known classes. This task is denoted as source-free universal domain adaptation (SF-UniDA) \cite{sf_unida}.

Interestingly, there exist some fields like few-shot learning \cite{few_shot} or class-incremental continual learning \cite{class_incremental} which adapt a pre-trained source model to new classes with a small labeled target dataset. The difference to the SF-UniDA task is that these methods are able to learn the new classes due to the available labeled target samples while SF-UniDA in an open-set or open-partial-set environment uses only unlabeled test-time samples and rejects samples of new classes as "unknown class". Furthermore, SF-UniDA jointly considers open-set recognition (category shift) and domain shift while the traditional few-shot and incremental-class continual learning methods do not consider a simultaneous domain shift. 

In addition to the ability to deal with domain and category shifts, many real-world applications require real-time inference of test data coming in as a stream. In other words, the test batches can only be accessed one at a time and each batch is only available once. Hence, each batch demands an immediate prediction and the TTA happens in parallel to inference. This is referred to as online TTA in contrast to offline TTA where an unlimited access to a finite test set is assumed and the whole adaptation process takes place before the prediction. Obviously, the online setting is more challenging and universally applicable to both online and offline scenarios.

Although some approaches consider the combination of domain and category shift \cite{universal_sf_da, kundu2020towards, umad, shot, feng2021open, sf_unida, li2023robustness}, we identified that there is a lack of methods combining all three paradigms online, source-free and universal, i.e. methods towards \emph{online} SF-UniDA. We overcome this lack by proposing the novel method Contrastive Mean Teacher (COMET) which is tailored to this challenging scenario. It leverages a mean teacher (MT) \cite{tarvainen2017mean, dobler2023robust} to get reliable pseudo-labels. Thereby, we follow the idea of \cite{feng2021open} and apply two entropy thresholds to be able to pseudo-label samples as unknown and to ignore samples with uncertain pseudo-labels to prevent negative transfer. Subsequently, we apply a contrastive loss \cite{khosla2020supervised} to adapt to the domain shift and enforce a feature space where not only the samples of the known classes form distinct clusters but also the unknown samples are clearly separated from these clusters. Additionally, we apply an entropy loss to ensure that the classifier output has a small entropy for samples of the known classes and a large entropy for unknown samples. This allows us to reject samples as unknown based on the entropy of the classifier output during the final prediction.

We evaluate our method on the two DA datasets DomainNet and VisDA-C. Thereby, COMET yields state-of-the-art results and shows its superiority over the competing methods by consistently performing well across all category shifts. We hope that our initial benchmark encourages future works to pay more attention to the realistic \emph{online} SF-UniDA scenario.

We summarize our contributions as follows:
\begin{itemize}
	\item We are the first to study the realistic but challenging task of \emph{online} SF-UniDA.
	\item We propose COMET, a method tackling this difficult task by applying a combination of contrastive learning and entropy optimization embedded into a mean teacher.
	\item We extensively evaluate COMET and create an initial baseline for \emph{online} SF-UniDA.
	\item We conduct rich ablation studies to investigate the impact of various design choices.
\end{itemize}

\section{Related work}
\subsection{Universal domain adaptation}
While the large majority of unsupervised domain adaptation methods is limited to the closed-set assumption, numerous approaches have been proposed in the recent years tackling either partial-set domain adaptation (PDA) \cite{cao2018partial, cao2018partial2, zhang2018importance}, open-set domain adaptation (ODA) \cite{open_set_domain_adaptation, saito2018open, psdc, liu2019separate, busto2018open, baktashmotlagh2018learning, bucci2020effectiveness} or open-partial-set domain adaptation (OPDA) \cite{you2019universal, fu2020learning, saito2021ovanet}. However, methods that perform well in one of these scenarios do not necessarily do so in the others. For example, \cite{you2019universal} showed that their OPDA method even underperforms the non-adapted source model for a PDA problem. In contrast, universal domain adaptation methods \cite{saito2020universal, Li_2021_CVPR, chen2022geometric} aim to perform well regardless of the category shift and without needing prior knowledge about its nature. Nevertheless, like all methods mentioned above, they require access to the source data which is not only inefficient but may also violate privacy or accessibility limitations.

\subsection{Online Test-time adaptation}
In contrast, TTA \cite{tta_survey} adapts a standard pre-trained source model to the target data without accessing the source data. Depending on whether the test data is available as a finite set with unlimited access or is received continuously as a stream of batches, one distinguishes between offline and online TTA. In this paper, we focus on online TTA, which is more realistic and universally applicable but also more challenging. In recent literature, mainly the calibration of the batch normalization (BN) statistics, self-training and entropy minimization have emerged as being most promising for this purpose. BN calibration methods \cite{tent, niu2023towards} root on the finding that adapting the BN statistics can significantly improve the performance on corrupted data \cite{NEURIPS2020_85690f81}. Self-training \cite{AdaContrast, shot, goyal2022test, dobler2023robust} uses the generation of pseudo-labels for the unlabeled target domain to guide the adaptation process. Finally, simply minimizing the entropy of the model output has also proved to be effective \cite{tent, shot, pmlr-v162-niu22a}. However, nearly all existing works just cope with a domain shift but do not consider the additional possibility of a category shift. 

\subsection{Source-free universal domain adaptation}
\label{sec:sfunida}
\begin{table}
	\caption{Overview whether the TTA approaches tackling both domain and category shift are universal, source-free and online}
	\label{tab:SF-UniDA}
	\vspace*{-0.1cm}
	\begin{tabularx}{\linewidth}{l*{8}{c}}
		\toprule
		 & \hspace*{-0.1cm}Ours & \hspace*{-0.1cm}\cite{sf_unida} & \hspace*{-0.1cm}\cite{universal_sf_da} & \hspace*{-0.1cm}\cite{kundu2020towards} & \hspace*{-0.1cm}\cite{umad} & \hspace*{-0.1cm}\cite{shot} & \hspace*{-0.1cm}\cite{feng2021open} & \hspace*{-0.1cm}\cite{li2023robustness}\\
		\midrule
		Universal & \cmark & \cmark & \cmark & \xmark & \cmark & \xmark & \xmark & \xmark \\
		Source-free & \cmark & \cmark & \xmark & \xmark & \xmark & \cmark & \cmark & \cmark \\
		Online & \cmark & \xmark & \xmark & \xmark & \xmark & \xmark & \xmark & \cmark \\
		\bottomrule
	\end{tabularx}
	\vspace*{-0.3cm}
\end{table}
Test-time adaptation to both domain and category shift is so far only addressed by \cite{universal_sf_da, kundu2020towards, umad, shot, feng2021open, sf_unida, li2023robustness}. Table \ref{tab:SF-UniDA} gives an overview of whether these methods fulfill the three paradigms universal, source-free and online. \cite{universal_sf_da, kundu2020towards, umad} build upon a dedicated open-set source training that already prepares the source model for the two shifts. In this way, they are limited in their practical application and from our understanding not truly source-free because they use more source knowledge than just a standard pre-trained source model. Among the remaining methods only \cite{sf_unida} is universal while \cite{feng2021open} and \cite{li2023robustness} are only designed for ODA and \cite{shot} needs prior knowledge about the nature of the category shift to be adapted accordingly. Moreover, of all these methods, only \cite{li2023robustness} is designed for the online scenario. In the following, we will take a closer look at the only two methods \cite{sf_unida, li2023robustness} lacking only one of the three paradigms online, source-free and universal.

\cite{sf_unida} proposes global and local clustering (GLC), a pseudo-labeling technique which is specially designed to cope with a category shift. Subsequently, they apply a cross-entropy loss for domain adaptation and an entropy threshold to reject unknown samples during inference. However, we doubt that their pseudo-labeling can still perform well when applied in an online manner because their clustering and kNN-based technique requires more data than just one batch as we will discuss in section \ref{pseudo_labeling}.

Finally, \cite{li2023robustness} most recently introduced an online TTA method being robust to open-world scenarios. They apply a contrastive loss which minimizes the distances between the samples and their respective most similar prototype within a prototype pool. This pool is initialized with the source prototypes but dynamically extended by prototypes representing the unknown class. Additionally, they apply distribution alignment. However, the open-world scenario they consider is neither universal nor what is commonly regarded as open-set. Instead, they pollute the test data with strong out-of-distribution samples from different datasets or even with random noise. In contrast, an open-set category shift normally adds samples of classes of the same dataset which were withheld during source training but have the same overall appearance as the samples of known classes. Hence, their method still has to prove whether it can also perform well in an online SF-UniDA scenario.

\begin{figure*}[!t]
	\centering
	\includegraphics[width=0.95\linewidth]{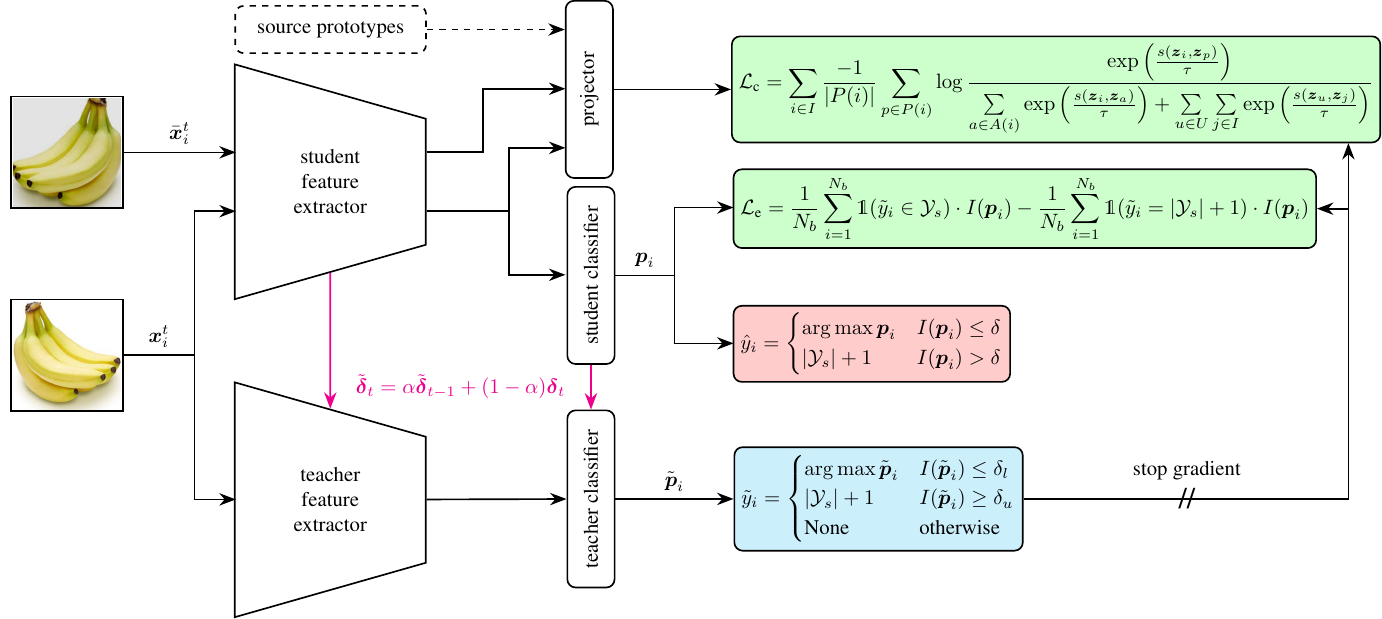}
	\caption{Overview of COMET. Both the student and teacher are initialized with the weights of the source model. Subsequently, for each target batch $\{x_i^t\}_{i=1}^{N_b}$ first the predictions $\hat{y}_i$ are generated using the output of the student classifier $\boldsymbol{p}_i$ (red box). Second, the pseudo-labels are generated using the classifier output of the teacher $\tilde{\boldsymbol{p}}_i$ (blue box). Third, the pseudo-labels are used to calculate both the contrastive loss $\mathcal{L}_c$ and the entropy loss $\mathcal{L}_e$ (green boxes) to update the student model by backpropagation. Lastly, the teacher weights are updated as an exponential moving average of the student weights (magenta).}
	\label{fig:framework}
	\vspace*{-0.3cm}
\end{figure*}

\section{Method}
\subsection{Preliminaries}
As the name suggests, the task of online SF-UniDA is to adapt a pre-trained source model to a target domain under the three paradigms online, source-free and universal. ''Source-free`` means that no source data can be leveraged for the adaptation due to privacy, memory or accessibility constraints. Using our strict understanding of this term to prevent any limitations of our method, we additionally assume that we cannot influence the source training. Hence, we only have access to a source model $f_s=h_s\circ g_s$, consisting of a feature extractor $g_s$ and a classifier $h_s$, which is pre-trained on the source dataset $\mathcal{D}_s=\{\mathcal{X}_s, \mathcal{Y}_s\}$ in a standard closed-set manner.

The goal of the adaptation process is to enable the model $f_s$ to classify the unlabeled target dataset $\mathcal{D}_t=\{\mathcal{X}_t\}$ as accurately as possible. Thereby, we assume that besides a domain shift $\mathcal{D}_t$ can also be subject to a category shift meaning that its label space $\mathcal{Y}_t$ may differ from $\mathcal{Y}_s$. Again, this includes three cases, namely partial-set ($\mathcal{Y}_t\subset \mathcal{Y}_s$), open-set ($\mathcal{Y}_s\subset \mathcal{Y}_t$) and open-partial-set ($\mathcal{Y}_s\cap\mathcal{Y}_t\neq\emptyset$, $\mathcal{Y}_s\nsubseteq\mathcal{Y}_t$, $\mathcal{Y}_s\nsupseteq\mathcal{Y}_t$) domain adaptation. Since in practice there is no prior knowledge about the category shift, the adaptation method needs to be universal by performing well for all these scenarios. Accordingly, it must reliably reject samples of new classes as unknown while correctly classifying the samples of the known classes.

Additionally, this paper focuses on the realistic online setting where each target batch $\{x_i^t\}_{i=1}^{N_b}$ can only be accessed once and requires an immediate prediction. This is difficult since the adaptation must be performed on the fly in parallel to inference which implies that for each adaptation step only the data of the current batch can be used.

In the following, we introduce our novel method Contrastive Mean Teacher (COMET) tackling online SF-UniDA. First, we discuss the challenges of pseudo-labeling in the online context and propose a technique tailored to the online and universal setting. Second, we describe the contrastive loss we use to adapt to the domain shift. Finally, we introduce the entropy loss which complements the contrastive loss and allows us to reject samples of unknown classes based on an entropy threshold. Figure \ref{fig:framework} shows an overview of our method.


\subsection{Pseudo-labeling}
\label{pseudo_labeling}
Self-training is a widely applied paradigm for TTA and proved to be effective. Thereby, reliable pseudo-labels are crucial for a successful adaptation and to prevent negative transfer. Most existing pseudo-labeling techniques (e.g. \cite{shot, sf_unida, AdaContrast}) leverage clustering or kNN-based methods. However, both require a large amount of samples to represent the underlying data distribution as well as possible in order to provide accurate pseudo-labels. This requirement is typically not fulfilled in online TTA since we only have access to one batch of data at each time. Accordingly, the direct application of clustering or kNN-based pseudo-labeling does not make sense for the online scenario.

To overcome this limitation, we introduce a pseudo-labeling tailored to the online setting. The basic idea is adopted from the simple yet effective pseudo-labeling of \cite{feng2021open}. Concretely, we directly work with the softmax classifier output and use its normalized entropy as an indicator of the confidence of the predicted label. A confident prediction will lead to a small entropy while uncertain predictions are characterized by a large entropy. Nevertheless, in the presence of a category shift, highly uncertain predictions are likely for samples of new classes and can therefore be pseudo-labeled as unknown with a large confidence. Since wrong pseudo-labels lead to a negative transfer, our intuition is that the quality of the pseudo-labels is more important than the quantity of pseudo-labeled samples used for the adaptation. Accordingly, we only want to use the samples with the most confident pseudo-labels for the adaptation. Therefore, we introduce two thresholds $\delta_l$ and $\delta_u$ on the entropy to divide the samples of each batch into three categories: samples with confident pseudo-labels of the known classes, samples confidently pseudo-labeled as unknown with the class index $|\mathcal{Y}_s|+1$ and uncertain samples. Hence, we assign the pseudo-labels to the samples of each target batch $\{\boldsymbol{x}_i^t\}_{i=1}^{N_b}$ as follows:
\begin{align}
	\tilde{y}_i = \begin{cases}
		\arg\max \tilde{\boldsymbol{p}}_i & I(\tilde{\boldsymbol{p}}_i)\leq \delta_l\\
		|\mathcal{Y}_s|+1 & I(\tilde{\boldsymbol{p}}_i)\geq \delta_u\\
		\text{None} & \text{otherwise}
	\end{cases}
	\label{eq:pseudo-label}
\end{align}
where 
\begin{align}
	I(\tilde{\boldsymbol{p}}_i)=-\frac{1}{\log |\mathcal{Y}_s|}\cdot\tilde{\boldsymbol{p}}_i^T\cdot\log \tilde{\boldsymbol{p}}_i
	\label{eq:entropy}
\end{align}
denotes the normalized entropy of the classifier output $\tilde{\boldsymbol{p}}_i$. The normalization by $\log |\mathcal{Y}_s|$ ensures that $I(\tilde{\boldsymbol{p}}_i)$ ranges only within the unit interval $[0,1]$. Empirically, we found that choosing the two thresholds symmetrically, i.e. $\delta_u=1-\delta_l$, is in general a good choice. In this way, the two thresholds are defined by only one hyperparameter which can e.g. be chosen such that a certain percentage of samples is left out in each target batch.

To further increase the robustness and quality of the pseudo-labeling, we embed this basic idea into a mean teacher (MT) framework \cite{tarvainen2017mean, dobler2023robust}. Thereby, both the student and the teacher model are initialized with the weights of the pre-trained source model. Subsequently, the student model is trained by backpropagation using the loss functions we will introduce in the following subsections while the non-trainable weights of the teacher are updated by an exponential moving average
\begin{align}
	\label{eq:RMT}
	\tilde{\boldsymbol{\delta}}_t = \alpha\tilde{\boldsymbol{\delta}}_{t-1}+(1-\alpha)\boldsymbol{\delta}_t~.
\end{align}
Here, $\tilde{\boldsymbol{\delta}}_t$ and $\tilde{\boldsymbol{\delta}}_{t-1}$ denote the weights of the teacher at the time instances $t$ and $t-1$, respectively, and $\boldsymbol{\delta}_t$ represents the weights of the student at time instance $t$. Furthermore, $\alpha$ is the momentum factor controlling the updating speed of the teacher. By averaging the weights of the student, the teacher's predictions become more robust because they are less prone to the short-term changes caused by the adaptation process. Therefore, we use the classifier output of the teacher model $\tilde{\boldsymbol{p}}_i$ to generate the pseudo-labels according to Eq. \ref{eq:pseudo-label} and \ref{eq:entropy}.

\subsection{Contrastive loss}
The basic idea of contrastive learning \cite{chen2020simple} is to learn a meaningful feature space by minimizing the distances within so-called positive data pairs while at the same time the distances within negative pairs are maximized. Originally designed for self-supervised learning, it has already been successfully adapted not only to supervised learning \cite{khosla2020supervised} but also to online TTA \cite{AdaContrast, dobler2023robust} and online open-world TTA \cite{li2023robustness}.

It is commonly agreed that learning a consistent, meaningful and preferably domain invariant feature space in which the classes are separable is key for any classification task. After a successful pre-training of the source model, this should be satisfied and we assume the source data to form distinct clusters in the feature space. However, a domain shift causes these clusters to diffuse and blur. Moreover, due to the category shift, samples of new classes can occur which do not belong to any of the original clusters.

We believe that the main task of online SF-UniDA is to restore the original state of the feature space where not only the samples of the known classes form distinct clusters but also the samples of new classes can be clearly separated from these clusters. Thus, we decide to work with a contrastive loss since it not only already proved to be a valuable tool for TTA but also allows us to directly enforce the desired properties on the feature space. To achieve the rebuilding process of the distinct clusters and the separation of the samples of new classes, we design our contrastive loss to minimize the following distances between the pseudo-labeled samples of each target batch:
\begin{itemize}
	\item distances among the samples with the same pseudo-label of a known class
	\item distance between each sample pseudo-labeled as one of the known classes and its corresponding class prototype
\end{itemize}
Moreover, it maximizes the following distances between the pseudo-labeled samples of each target batch:
\begin{itemize}
	\item distances between the samples with different pseudo-labels (including the distances between the samples pseudo-labeled as one of the known classes and the samples pseudo-labeled as unknown)
	\item distances between the pseudo-labeled samples and the class prototypes which do not correspond to their respective pseudo-label (i.e. for the samples pseudo-labeled as unknown, the distances to all prototypes are maximized)
\end{itemize}
Thereby, we extend the set of pseudo-labeled samples by one augmentation of each sample. In this way, i.e. by enforcing a small distance between the samples and their respective augmentation, a consistent feature space is learned which is invariant to small changes in the input space and therefore more robust to domain shifts.

Our contrastive loss is computed for each target batch $X_n^t = \{\boldsymbol{x}_i^t\}_{i=1}^{N_b}$ by first extracting the samples pseudo-labeled as known classes $X_n^k = \{\boldsymbol{x}_i^t\in X_n^t| \tilde{y}_i\in\mathcal{Y}_s\}$ and the samples pseudo-labeled as unknown $X_n^u = \{\boldsymbol{x}_i^t\in X_n^t| \tilde{y}_i=|\mathcal{Y}_s|+1\}$. Let $N_n^k$ and $N_n^u$ denote the number of samples in $X_n^k$ and $X_n^u$, respectively. Second, we extend both sets $X_n^k$ and $X_n^u$ by adding one augmentation $\bar{\boldsymbol{x}}_i^{t}$ of each sample $\boldsymbol{x}_i^t$ and use the student feature extractor $g(\cdot)$ to transform everything into the feature space resulting in the two sets $R_n^k=\{g(\boldsymbol{x}_i^t)|\boldsymbol{x}_i^t\in X_n^k\}$ and $R_n^u=\{g(\boldsymbol{x}_i^t)|\boldsymbol{x}_i^t\in X_n^u\}$. Subsequently, we further extend each sample-augmentation pair $(g(\boldsymbol{x}_i^t), g(\bar{\boldsymbol{x}}_i^t))$ within the set $R_n^k$ by adding the class prototype that matches their respective pseudo-label. In a final preparation step, we use a multi-layer perceptron with a single hidden layer $Proj(\cdot)$ to map all feature representations within $R_n^k$ and $R_n^u$ into a lower dimensional projection space as proposed by \cite{khosla2020supervised}. This results in the set $\{\boldsymbol{z}_i\}_{i=1}^{3N_n^k+2N_n^u}=\{Proj(\boldsymbol{r}_i)|\boldsymbol{r}_i\in R_n^k\cup R_n^u\}$, where the first $3N_n^k$ elements, i.e. the elements whose indices are within the index set $I^k=\{1,\ldots, 3N_n^k\}$, correspond to the sample-augmentation-prototype triples pseudo-labeled as a known class. Accordingly, the final $2N_n^u$ elements whose indices are contained in the index set $I^u=\{3N_n^k+1,\dots, 3N_n^k+2N_n^u\}$ correspond to the sample-augmentation pairs pseudo-labeled as unknown. Finally, we compute the contrastive loss as
\begin{equation}
	\mathcal{L}_\text{c}=\sum_{i\in I^k}\frac{-1}{|P(i)|}\hspace{-0.095cm}\sum_{p\in P(i)}\hspace{-0.12cm}\log \frac{\exp\left(\frac{s(\boldsymbol{z}_i,\boldsymbol{z}_p)}{\tau}\right)}{\left(\splitdfrac{\sum\limits_{a\in A(i)}\exp\left(\frac{s(\boldsymbol{z}_i,\boldsymbol{z}_a)}{\tau}\right)} {\hspace{-0.6cm}+\hspace{-0.15cm}\sum\limits_{u\in I^u}\sum\limits_{j\in I^k}\exp\left(\frac{s(\boldsymbol{z}_u,\boldsymbol{z}_j)}{\tau}\right)}%
		\right)}
\end{equation}
where $A(i)=I^k\backslash \{i\}$ and $P(i)=\{p\in A(i)|\tilde{y}_p=\tilde{y}_i\}$, i.e. $P(i)$ denotes the index set for all samples, augmentations and prototypes with the same pseudo-label as the element at index $i$. Moreover, $\mathrm{s}(\boldsymbol{z}_a,\boldsymbol{z}_b)=\boldsymbol{z}_a^T\boldsymbol{z}_b/(||\boldsymbol{z}_a||\,||\boldsymbol{z}_b||)$ is the cosine similarity and $\tau$ denotes the temperature.

\subsection{Source-free prototypes of known classes}
As mentioned above, our contrastive loss roots on prototypes in the feature space for all known classes. They serve as cluster centers to gather the samples with the corresponding pseudo-labels as closely as possible around them. The existing contrastive learning-based TTA methods \cite{dobler2023robust, li2023robustness} use so-called source prototypes, i.e. the class-wise means of the feature representations of the source data after pre-training. In this way, the clusters in the target domain stay at the same positions as in the source domain and therefore the classifier does not require much adaptation. Although the usage of such source prototypes is commonly considered as ''source-free`` since it neither violates privacy nor efficiency constraints and also requires only little memory space, we argue that in a strictly source-free setting where the source training cannot be influenced the availability of source prototypes cannot be guaranteed. Accordingly, we propose a strictly source-free variation to acquire class prototypes for our contrastive loss without any information from the source training (e.g. source prototypes) except for the pre-trained source model $f_s$. Thereby, we continuously calculate the sums of the feature representation of all test-time samples with the same pseudo-labels across all processed target batches while keeping track of the numbers of samples which contributed to the individual sums. In this way, we can dynamically calculate the class-wise means in the target domain at each time instance and use them as strictly source-free class prototypes. In the remainder of this paper, we will refer to the variation of our method using these strictly source-free prototypes as COMET-F while we will denote the version using source prototypes as COMET-P.

\subsection{Entropy loss}
Since the contrastive loss solely works with the feature extractor to restore the desired properties of the feature space, a complementary loss is helpful to also update the classifier in parallel. The goal is to enable it to better translate the adapted target feature representations into the class predictions. Ultimately, each sample of a known class should be correctly predicted with a large confidence while the unknown samples should lead to a uniformly distributed softmax output such that we can reject them confidently based on an entropy threshold. Our intuition is that the latter, i.e. yielding an output with a large entropy for the samples of new classes, still needs to be incorporated into the classifier while the first is already mainly satisfied. This is because the contrastive loss enforces the target samples of the known classes to cluster in the same areas in the feature space as the source samples, especially if source prototypes are used. However, during the closed-set source training, the classifier was not taught how to deal with samples in the space between these areas and without adaptation we cannot guarantee that its corresponding output will have a high entropy.

Therefore, we apply an additional entropy loss to enable a reliable differentiation between target samples of known and unknown classes based on the entropy of the classifier output. An entropy-based loss not only already proved to be effective for TTA \cite{tent, shot, pmlr-v162-niu22a} but also allows us to directly enforce the desired behavior. Concretely, we minimize the entropy of the softmax classifier output for the samples pseudo-labeled as a known class and maximize the entropy for the samples pseudo-labeled as unknown. This is achieved by minimizing the following loss function for each target batch $\{\boldsymbol{x}_i^t\}_{i=1}^{N_b}$:
\begin{equation}
\begin{aligned}
	\mathcal{L}_\text{e}=&\frac{1}{N_b}\sum_{i=1}^{N_b}\mathds{1}(\tilde{y}_i\in\mathcal{Y}_s)\cdot I(\boldsymbol{p}_i)\\
	- &\frac{1}{N_b}\sum_{i=1}^{N_b} \mathds{1}(\tilde{y}_i=|\mathcal{Y}_s|+1)\cdot I(\boldsymbol{p}_i)~.
\end{aligned}
\end{equation}
Thereby, $\mathds{1}()$ is the indicator function, i.e. it is one if the condition in the parentheses is fulfilled and zero otherwise, and $I(\boldsymbol{p}_i)$ denotes the normalized entropy of the classifier output $\boldsymbol{p}_i$ of the student model. The calculation of $I(\boldsymbol{p}_i)$ is equivalent to the calculation w.r.t. the classifier output $\tilde{\boldsymbol{p}}_i$ of the teacher model given in Eq. \ref{eq:entropy}.

The overall loss function to update the student model is given as
\begin{align}
	\mathcal{L}= \mathcal{L}_\text{c} + \lambda \mathcal{L}_\text{e}
\end{align}
where $\lambda>0$ serves to balance the two loss functions.

\subsection{Inference}
For the inference of each target batch $\{\boldsymbol{x}_i^t\}_{i=1}^{N_b}$ in parallel to the adaptation, we use the classifier output of the student model $\boldsymbol{p}_i$. For each sample $\boldsymbol{x}_i^t$, we first compute its entropy $I(\boldsymbol{p}_i)$ as defined in Eq. \ref{eq:entropy} and compare it to a threshold $\delta$. On this basis, we either classify it according to the maximum of $\boldsymbol{p}_i$ or reject it as unknown as shown in the following:
\begin{align}
	\label{eq:inference}
	\hat{y}_i = \begin{cases}
		\arg\max \boldsymbol{p}_i & I(\boldsymbol{p}_i)\leq \delta\\
		|\mathcal{Y}_s|+1 & I(\boldsymbol{p}_i)> \delta\end{cases}~.
\end{align}
Note that we use only one entropy threshold here to assign a label $\hat{y}_i$ to all samples $\boldsymbol{x}_i^t$ in contrast to our pseudo-labeling (Eq. \ref{eq:pseudo-label}) where we use two thresholds to leave samples with uncertain predictions out of the adaptation process. Moreover, while we use the teacher model for pseudo-labeling, we use the student model for inference because it adapts faster to the domain shift. This is important since in the online scenario the inference needs to happen in parallel to the adaptation.

\section{Experiments}
\subsection{Setup}
We mainly resort to the setup of \cite{sf_unida}. Accordingly, we use a classical TTA setup instead of continual TTA, meaning we consider the adaptation to a single domain shift without requiring the preservation of the performance on the source domain. However, we strictly follow an \emph{online} TTA setting where only one forward pass is performed for each batch.
\subsubsection{Datasets}
\begin{table}
	\caption{Class splits of the datasets, i.e. number of the shared, source-private and target-private classes, for the three category shift scenarios OPDA, ODA and PDA, respectively}
	\label{tab:class_splits}
	\vspace*{-0.1cm}
	\begin{tabularx}{\linewidth}{l*{3}{c}}
		\toprule
		& \multicolumn{3}{c}{$|\mathcal{Y}_s\cap\mathcal{Y}_t|$, $|\mathcal{Y}_s\backslash\mathcal{Y}_t|$, $|\mathcal{Y}_t\backslash\mathcal{Y}_s|$}\\
		\cmidrule(lr){2-4}
		& PDA & ODA & OPDA\\
		\midrule
		DomainNet \hspace{0.5cm} & 200, 145, 0 & 200, 0, 145 & 150, 50, 145 \\
		VisDA-C & 6, 6, 0 & 6, 0, 6 & 6, 3, 3 \\
		\bottomrule
	\end{tabularx}
	\vspace*{-0.3cm}
\end{table}
We evaluate COMET on the two DA datasets DomainNet \cite{domainnet} and VisDA-C \cite{visda}. DomainNet consists of about 0.6 million images of 345 classes across the three domains painting (P), real (R) and sketch (S). All 345 classes are available in all three domains. We will evaluate all six domain shifts P$\rightarrow$R, P$\rightarrow$S, R$\rightarrow$P, R$\rightarrow$S, S$\rightarrow$P and S$\rightarrow$R. VisDA-C consists of only 12 object classes but features a challenging domain shift from synthetically generated 2D renderings of 3D models to real-world images taken from the Microsoft COCO dataset \cite{Coco}. As usual in the literature, we use the first domain as the source domain containing 152,397 images while the second domain is the target domain consisting of 55,388 images.

To create the three category shift scenarios PDA, ODA and OPDA, we order the classes alphabetically and use the first $|\mathcal{Y}_s|$ of them as source classes and the last $|\mathcal{Y}_t|$ as target classes. Accordingly, $|\mathcal{Y}_s\cap\mathcal{Y}_t|$ classes are shared. The class splits, i.e. the number of shared, source-private and target-private classes, are given in Table \ref{tab:class_splits} for both datasets, respectively. In this way, domain shift and different kinds of category shifts happen simultaneously.

\subsubsection{Competing methods}
Since we are the first to study the combination of the three paradigms online, source-free and universal DA, there are yet no other methods designed for this scenario. Nevertheless, a fair comparison is only possible if the competing methods also fulfill these three properties. Concretely, since in the offline scenario the whole adaptation process takes place before the inference while in the online scenario both needs to happen in parallel, better results can naturally be achieved in the offline scenario. Therefore, a comparison of methods across both scenarios does not make sense. The same applies to the source-free paradigm, where methods naturally achieve better results if source data can be leveraged. Finally, we obviously require the competing methods to be universal because we want to evaluate our method across all category shifts PDA, ODA and OPDA.

The first obvious choice as a competing method is to apply the non-adapted source model and use an entropy threshold to reject samples as unknown like we do in COMET (see Eq. \ref{eq:inference}). This serves as a lower bound TTA methods need to outperform since otherwise they suffer from negative transfer. Second, we apply the GLC method \cite{sf_unida} batch-wise in the online scenario. Finally, we also apply the OWTTT method \cite{li2023robustness} although it was designed for an open-world TTA scenario which is neither universal nor what is commonly considered as ODA as explained at the end of subsection \ref{sec:sfunida}. We will evaluate whether it can still perform well for online SF-UniDA.

\begin{table*}
	\caption{Results of the experiments on DomainNet and VisDA-C}
	\label{tab:results}
	\vspace*{-4mm}
	\begin{subtable}[t]{\textwidth}
		\caption{Accuracy in $\%$ for the PDA scenario}
		\label{tab:results_PDA}
		\begin{tabularx}{\textwidth}{p{6cm} *{6}{c} a|*{1}{c}}
			\toprule
			PDA & P$\rightarrow$R & P$\rightarrow$S & R$\rightarrow$P & R$\rightarrow$S & S$\rightarrow$P & S$\rightarrow$R & Avg. & VisDA-C\\
			\midrule
			Source-only & $41.5$ & $24.0$ & $30.5$ & $22.7$ & $17.8$ & $25.2$ & $26.95$ & $17.1$ \\
			Online GLC \cite{sf_unida} & $16.6$ & $16.8$ & $12.8$ & $11.8$ & $16.0$ & $14.5$ & $14.75$ & $21.7$ \\
			OWTTT \cite{li2023robustness} & $37.2$ & $26.0$ & $29.1$ & $24.8$ & $24.5$ & $25.7$ & $27.88$ & $28.1$ \\
			COMET-P (Ours) & \boldmath$51.0$ & \boldmath$34.0$ & \boldmath$38.0$ & \boldmath$32.7$ & \boldmath$28.5$ & \boldmath$39.2$ & \boldmath$37.23$ & \boldmath$33.4$ \\
			COMET-F (Ours) & $48.5$ & $32.5$ & \boldmath$38.0$ & $31.9$ & $27.1$ & $36.9$ & $35.82$ & $33.0$ \\
			\bottomrule
		\end{tabularx}
	\end{subtable}
	\vspace*{0mm}
	
	\begin{subtable}[t]{\textwidth}
		\caption{H-score in $\%$ for the ODA scenario}
		\label{tab:results_ODA}
		\begin{tabularx}{\textwidth}{p{6cm} *{6}{c} a|*{1}{c}}
			\toprule
			ODA & P$\rightarrow$R & P$\rightarrow$S & R$\rightarrow$P & R$\rightarrow$S & S$\rightarrow$P & S$\rightarrow$R & Avg. & VisDA-C\\
			\midrule
			Source-only & $57.8$ & $40.4$ & $47.0$ & $38.4$ & $30.7$ & $45.9$ & $43.37$ & $31.7$ \\
			Online GLC \cite{sf_unida} & $41.3$ & $29.0$ & $23.1$ & $20.5$ & $32.6$ & $33.4$ & $29.98$ & $44.7$ \\
			OWTTT \cite{li2023robustness} & $54.7$ & $45.8$ & $46.5$ & $42.8$ & \boldmath$47.0$ & $53.2$ & $48.33$ & \boldmath$56.0$ \\
			COMET-P (Ours) & \boldmath$58.5$ & \boldmath$47.8$ & \boldmath$51.5$ & \boldmath$46.5$ & $39.0$ & \boldmath$53.3$ & \boldmath$49.43$ & $53.3$ \\
			COMET-F (Ours) & $58.2$ & $47.3$ & $50.9$ & $45.9$ & $37.6$ & $52.1$ & $48.67$ & $52.7$ \\
			\bottomrule
		\end{tabularx}
		
	\end{subtable}
	\vspace*{0mm}
	
	\begin{subtable}[t]{\textwidth}
		\caption{H-score in $\%$ for the OPDA scenario}
		\label{tab:results_OPDA}
		\begin{tabularx}{\textwidth}{p{6cm} *{6}{c} a|*{1}{c}}
			\toprule
			OPDA & P$\rightarrow$R & P$\rightarrow$S & R$\rightarrow$P & R$\rightarrow$S & S$\rightarrow$P & S$\rightarrow$R & Avg. & VisDA-C\\
			\midrule
			Source-only & $57.6$ & $38.8$ & $47.6$ & $38.6$ & $32.2$ & $47.9$ & $43.78$ & $26.9$ \\
			Online GLC \cite{sf_unida} & $34.0$ & $29.6$ & $27.3$ & $23.0$ & $34.2$ & $38.0$ & $31.02$ & $39.1$ \\
			OWTTT \cite{li2023robustness} & $56.3$ & $45.4$ & $47.8$ & $42.9$ & \boldmath$48.2$ & $53.6$ & $49.03$ & \boldmath$47.3$ \\
			COMET-P (Ours) & \boldmath$58.9$ & \boldmath$46.6$ & \boldmath$52.5$ & \boldmath$46.4$ & $39.2$ & \boldmath$54.3$ & \boldmath$49.65$ & $46.7$ \\
			COMET-F (Ours) & $58.8$ & $45.7$ & $51.8$ & $45.8$ & $38.3$ & $53.5$ & $48.98$ & $45.7$ \\
			\bottomrule
		\end{tabularx}
		
	\end{subtable}
	\vspace*{-0.3cm}
\end{table*}

\subsubsection{Implementation details}
For our source model, we adopt the same network structure and training protocol as \cite{shot} and \cite{sf_unida}. Concretely, we use a ResNet-50 \cite{he2016deep} pre-trained on ImageNet \cite{deng2009imagenet} with a $256$-dimensional feature space for both the student and teacher model and for both datasets. Moreover, we also use this architecture as the backbone of the competing methods to ensure a fair comparison. In the target domain, we work with a batch size of $128$ and apply a SGD optimizer with momentum $0.9$ and a learning rate of $0.0001$ for DomainNet and $0.001$ for VisDA-C. Furthermore, regarding our pseudo-labeling, we set the momentum for updating the mean teacher to $\alpha=0.999$ and the two entropy thresholds to $\delta_l=0.25$ and $\delta_u=0.75$. We choose the loss balancing factor to $\lambda=0.1$ for DomainNet and $\lambda=0.01$ for VisDA-C. Regarding the contrastive loss, we set the temperature to $\tau=0.1$ and the dimensions of the projection space to $128$. Finally, for inference, we use an entropy threshold of $\delta=0.5$ to reject samples as unknown.

\subsubsection{Evaluation metrics}
We stick to the common metrics used in the literature. Specifically, we report the classification accuracy over all target samples for PDA and the H-score for ODA and OPDA. The H-score denotes the harmonic mean of the accuracy of known and unknown samples.

\subsection{Results}
The results are shown in Table \ref{tab:results}. First, we focus on DomainNet. Thereby, it is striking that GLC significantly underperforms the source-only model in nearly all settings. This matches our intuition we explained in subsection \ref{pseudo_labeling} that clustering- and kNN-based pseudo-labeling is not suitable for the online scenario. Especially kNN can obviously not work properly if the number of classes exceeds the batch size like it is the case here. OWTTT, on the other hand, achieves on average a significant improvement over the source-only results for ODA and OPDA. However, this is not the case for PDA where its average improvement is less than $\SI{1}{\percent}$. Although PDA may appear to be the simplest task among the three category shifts at first glance, it is challenging in the context of universal DA because methods tend to falsely reject samples as unknown due to the domain shift. OWTTT seems to be also subject to this problem. Moreover, it does not perform equally well for all domain shifts. While it shows particularly strong results for the S$\rightarrow$P domain shift, it struggles with the P$\rightarrow$R and R$\rightarrow$P domain shifts across all three category shifts. On five of the six corresponding experiments it even falls behind the source-only results. In contrast, both variations of our method COMET consistently improve the source-only performance in all experiments. Especially in the PDA scenario both COMET-P and COMET-F significantly outperform all competing methods across all domain shifts. Nonetheless, they also show good results for ODA and OPDA. Concretely, COMET-P is able to achieve the best results for five of the six domain shifts for ODA and OPDA, respectively, and in this way also yields the largest average performance across all three category shifts. Not surprisingly, COMET-F overall performs a little worse than COMET-P. However, the margin between the two variations is only less than $\SI{1.5}{\percent}$ on average for all category shifts. In this way, COMET-F also outperforms OWTTT on average for PDA and ODA while both yield a nearly equal average performance for OPDA. This is remarkable since OWTTT relies on source prototypes while COMET-F does not.

Finally, we look at the results on VisDA-C. Although GLC seems to benefit from the small number of classes and can achieve improvements w.r.t. the source-only model for all category shifts, it is still clearly outperformed by the other three methods across all PDA, ODA and OPDA. In contrast to DomainNet, here OWTTT yields the best results for both ODA and OPDA. It seems to benefit more from the reduced number of classes than COMET. However, COMET-P is not far behind especially for OPDA. Moreover, both COMET variations still clearly outperform OWTTT for PDA. Like for DomainNet, COMET-F again performs only slightly, i.e. less than $\SI{1}{\percent}$, worse compared to COMET-P across all three category shifts.

\begin{table*}[t!]
	\caption{H-scores in $\%$ for different choices of the main hyperparameters, namely the momentum $\alpha$, the entropy threshold for inference $\delta$ and the entropy thresholds for pseudo-labeling $\delta_l$ and $\delta_u$, evaluated for the R$\rightarrow$S OPDA scenario.}
	\label{tab:ablation}
	\vspace*{-4mm}
	
	\begin{subtable}[t]{0.5\textwidth}
		\caption{Momentum factor $\alpha$}
		\begin{tabularx}{\linewidth}{l *{5}{c}}
			\toprule
			$\alpha$ & $0.99$ & $0.995$ & $0.999$ & $0.9995$ & $0.9999$ \\
			\midrule			
			COMET-P & \boldmath$46.4$ & \boldmath$46.4$ & \boldmath$46.4$ & $46.3$ & $46.2$\\
			COMET-F & \boldmath$45.8$ & $45.7$ & \boldmath$45.8$ & \boldmath$45.8$ & $45.7$\\
			\bottomrule
		\end{tabularx}
	\end{subtable}\hspace*{1cm}
	\begin{subtable}[t]{0.45\textwidth}
		\caption{Entropy threshold $\delta$ for inference}
		\begin{tabularx}{\linewidth}{l *{5}{c}}
			\toprule
			$\delta$ & $0.4$ & $0.45$ & $0.5$ & $0.55$ & $0.6$ \\
			\midrule
			COMET-P & $45.1$ & $46.1$ & $46.4$ & \boldmath$46.5$ & $46.2$\\
			COMET-F & $44.7$ & $45.5$ & $45.8$ & \boldmath$46.2$ & $46.0$\\
			\bottomrule
		\end{tabularx}
	\end{subtable}\vspace{0.3cm}

	\hspace{3.5cm}
	\begin{subtable}[t]{0.6\textwidth}
		\caption{Entropy thresholds $\delta_l$ and $\delta_u$ for pseudo-labeling}
		\begin{tabularx}{\linewidth}{l *{5}{c}}
			\toprule
			$\delta_l$/$\delta_u$ & $0.15$/$0.85$ & $0.2$/$0.8$ & $0.25$/$0.75$ & $0.3$/$0.7$ & $0.35$/$0.65$ \\
			\midrule
			COMET-P & $46.1$ & $46.3$ & \boldmath$46.4$ & \boldmath$46.4$ & $46.2$\\
			COMET-F & $45.4$ & $45.6$ & \boldmath$45.8$ & $45.6$ & $45.6$\\
			\bottomrule
		\end{tabularx}
	\end{subtable}
	\vspace*{-0.3cm}
\end{table*}

\subsection{Ablation studies}
For our ablation studies we use the R$\rightarrow$S OPDA scenario of DomainNet as a representative example. OPDA best demonstrates the trade-off between reliably rejecting samples of new classes and correctly classifying samples of known classes which is required for online SF-UniDA. The R$\rightarrow$S domain shift was chosen arbitrarily.
\subsubsection{Reduced batch sizes}
\begin{figure}[!t]
	\centering
	\includegraphics[width=0.9\linewidth]{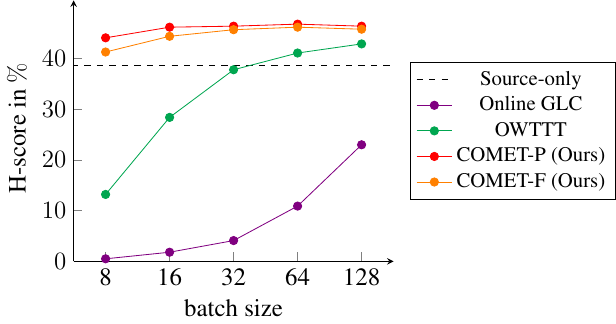}
	\caption{Performance development of all methods for reduced batch sizes shown for the R$\rightarrow$S OPDA scenario.}
	\label{fig:batch_size}
\end{figure}
\begin{figure}[t]
	\centering
	\includegraphics[width=0.9\linewidth]{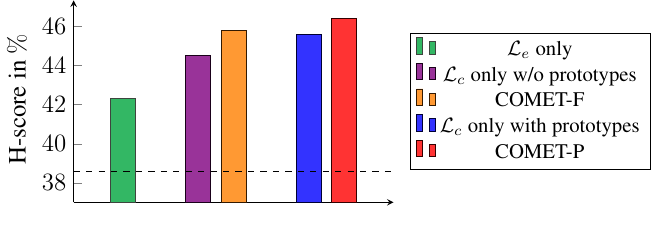}
	\caption{Results using different combinations of the losses $\mathcal{L}_c$ and $\mathcal{L}_e$ for the R$\rightarrow$S OPDA scenario. The dashed line indicates the source-only performance.}
	\label{fig:loss}
	\vspace*{-0.3cm}
\end{figure}
Figure \ref{fig:batch_size} shows how a reduction of the batch size influences the performance of the individual methods. Although GLC already underperforms the source-only model for the original batch size of $128$ it becomes even worse for smaller batch sizes and converges towards the performance of random guessing. Interestingly, also the performance of OWTTT decreases dramatically and already falls behind the source-only performance at a batch size of $32$. This may be due to their clustering-based determination of the rejection threshold. It seems like for smaller batch sizes there are not enough samples to represent the distribution of their outlier score and hence the rejection threshold cannot be chosen carefully. In contrast, COMET does not suffer from such a limitation and proves to be robust against a reduction of the batch size. The performance of COMET-P remains constant and eventually decreases only slightly for a batch size of $8$. Nevertheless, even in this case it still significantly improves the source-only performance. The same applies to COMET-F. Although its decline starts a bit earlier, namely at a batch size of $16$, it also still performs clearly better than the source-only model even for a batch size of only $8$.

\subsubsection{Contribution of each loss}
Figure \ref{fig:loss} shows how each of the two losses $\mathcal{L}_c$ and $\mathcal{L}_e$ contributes to the overall performance of COMET. It can be seen that both already significantly improve the source-only result when applied as stand-alone loss. Thereby, as expected, $\mathcal{L}_c$ both with and without using source prototypes performs better than $\mathcal{L}_e$ and therefore is responsible for the largest part of the success of COMET. Nonetheless, $\mathcal{L}_e$ is still a valuable addition and proves to complement the contrastive loss well since we clearly obtain the best results when both losses are combined.

\subsubsection{Hyperparameter sensitivity}
Table \ref{tab:ablation} shows the results that are obtained for different choices of the main hyperparameters of COMET, namely the momentum $\alpha$, the entropy threshold for inference $\delta$ and the entropy thresholds for pseudo-labeling $\delta_l$ and $\delta_u$. We observe that COMET maintains a stable performance in a broad range around the selected values for all of these hyperparameters. Therefore, their choice is not critical for the success of our method.

\section{Conclusion}
We identified online source-free universal domain adaptation (online SF-UniDA) as a so far overlooked task despite its great practical relevance. Therefore, we have proposed Contrastive Mean Teacher (COMET), a method that overcomes this limitation. It uses a contrastive loss to rebuild a feature space where samples of known classes build distinct clusters and samples of unknown classes are clearly separated from them. This is complemented by an entropy loss allowing us to reject the samples of new classes using an entropy threshold during inference. Both losses are embedded into a mean teacher framework to provide them with reliable pseudo-labels. Finally, our extensive experiments verified the effectiveness of COMET across all evaluated domain and category shifts. Future work could extend COMET to tackle continual TTA or to learn new classes in a zero-shot manner instead of only rejecting them.


\bibliographystyle{IEEEtran}

\bibliography{IEEEabrv, refs}

\end{document}